\documentclass[letterpaper, 10 pt, conference]{ieeeconf}  

\IEEEoverridecommandlockouts                              
\usepackage[pdftex]{graphicx}
\graphicspath{{../pdf/}{../png/}}
\usepackage{array}
\usepackage{mdwmath}
\usepackage{mdwtab}
\usepackage{eqparbox}
\usepackage{url}
\usepackage[noend]{algpseudocode}
\usepackage{algorithm}
\usepackage{mathtools}
\usepackage{booktabs}
\usepackage{cite}
\usepackage{algorithm,algpseudocode}
\usepackage{amsmath,amssymb,amsfonts}

\title{\LARGE \bf
Neural Motion Planning for Autonomous Parking
}

\author{Dongchan Kim and Kunsoo Huh

\thanks{Dongchan Kim and Kunsoo Huh are with the Department of Automotive Engineering, Hanyang University, Seoul, 04763, Republic of Korea
        {\tt\small {jookker, khuh2}@hanyang.ac.kr}}%
}

\begin{document}

\maketitle
\thispagestyle{empty}
\pagestyle{empty}

\begin{abstract}
This paper presents a hybrid motion planning strategy that combines a deep generative network with a conventional motion planning method. 
Existing planning methods such as A* and Hybrid A* are widely used in path planning tasks because of their ability to determine feasible paths even in complex environments; however, they have limitations in terms of efficiency.
To overcome these limitations, a path planning algorithm based on a neural network, namely the neural Hybrid A*, is introduced. 
This paper proposes using a conditional variational autoencoder (CVAE) to guide the search algorithm by exploiting the ability of CVAE to learn information about the planning space given the information of the parking environment.
An efficient expansion strategy is utilized based on a distribution of feasible trajectories learned in the demonstrations. 
The proposed method effectively learns the representations of a given state, and shows improvement in terms of algorithm performance.
\end{abstract}

\section{INTRODUCTION}
Recently, autonomous driving technology has gained popularity. 
Parking is an essential task in autonomous driving. 
The autonomous vehicle should recognize the surrounding environment, including the obstacles and free space, and reach the desired goal state while avoiding collisions.
Presently, autonomous parking systems are commercially available. In addition, autonomous valet parking that allows an autonomous vehicle to park the car is also available \cite{khalid2020smart}.

To date, many path planning algorithms have been developed, such as the artificial potential field (APF) \cite{dong2016experimental}, the rapidly exploring random tree (RRT) algorithm \cite{kuwata2008motion}, the RRT* algorithm \cite{vlasak2020accelerated}, the partial motion planning (PMP) algorithm \cite{benenson2006integrating}, the A* algorithm \cite{cheng2014improved} and the Hybrid A* algorithm \cite{dolgov2010path}.
The potential field method uses the APF-based collision-free holonomic path first and then uses the generated path for further optimization.

Then, the remainder algorithms can be divided into two approaches \cite{sharma2021recent}: sampling based methods and searching based methods. The sampling based methods include two approaches which consist of random sampling based methods and deterministic sampling based methods.
First, RRT and RRT* algorithms exist in the random sampling based methods.
The RRT-based method constructs trees to connect the nodes obtained from the given sampling distributions. A feasible path is obtained by traversing the tree through nodes.
The RRT* algorithm is used to accelerate the computational time by utilizing the \textit{rewiring} procedure.
Then, the PMP method represents for the deterministic sampling based method \cite{benenson2006integrating}.
This method performs local searching rather than searching entire space to reduce computational burden.

The searching based methods include heuristic based planning and state lattice based planning methods.
First, A* and Hybrid A* algorithms are included in heuristic based methods.
The A*-based method uses the A* algorithm to efficiently search for parking spots and serves as a large parking guidance system. It adopts a heuristic function to obtain the optimal path. 
The Hybrid A*-based method is used to search for a continuous state, and thereby it mitigates the limitation of classical A*, where only the centers of the grid can be visited.
The lattice based planning methods utilize the discretization of configuration space into a set of states \cite{bicchi2002reachability, pivtoraiko2005efficient}.
The lattice provides a solution as graph search in the motion planning problem.

Among the aforementioned algorithms, we focus on the application of the A* and Hybrid A* algorithms which are widely used for autonomous driving tasks.   
In particular, the Hybrid A* can generate kinodynamic paths in clustered environments using a simplified vehicle model \cite{dolgov2010path}, and the designed paths are close to the human driving style.
Therefore, the Hybrid A* algorithm is applied in our study. However, the Hybrid A* algorithm tends to become highly time and memory-intensive as the size of the configuration space grows. 

To overcome the limitations of the Hybrid A* algorithm, the strategy in searching which is a state expansion procedure should be improved. Inspired by existing studies on applying deep neural networks in planning problems \cite{ichter2018learning, wang2020neural}, a path planning algorithm with the help of a guidance map, which is a learned distribution of feasible trajectories, is utilized in this study.

Recently, deep generative networks have been actively studied.   
For example, a variational autoencoder (VAE), which is a popular method for learning a generative model of a set of data, was utilized \cite{chen2016dynamic}.
The VAE enables the representation of high-dimensional movements in a low-dimensional latent space. In addition, an optimal movement is available to reproduce the movements.
The conditional variational autoencoder (CVAE) incorporates additional conditional input to the VAE method \cite{sohn2015learning}.
The CVAE is a deep conditional generative model for structured output prediction that uses Gaussian latent variables. The input observations modulate the prior on the Gaussian latent variables that generate the output.
There are several studies on trajectory prediction and multi-modal prediction using the CVAE network where the additional condition could enhance the performance \cite{feng2019vehicle, bhattacharyya2019conditional}.

In this study, a CVAE network-based Hybrid A* algorithm is proposed. The CVAE model is trained to provide the predicted distribution of feasible trajectories when the parking environment, along with the initial and goal states, is given.
The contributions of this study can be summarized as follows:
\begin{itemize}
    \item A neural Hybrid A* algorithm for autonomous parking is proposed that combines a deep generative network and a conventional planning method. 
    \item The CVAE architecture is utilized to learn the feasible trajectory distribution given the map information including the initial and goal states and obstacles.
    \item The proposed method significantly reduces the computational time and number of nodes in the test scenarios.
    \item The feasibility of the proposed method is demonstrated in various autonomous parking scenarios in the simulation.
\end{itemize}

The remainder of this paper proceeds as follows. In Section 2, the conventional Hybrid A* algorithm is described in detail. 
In Section 3, the neural Hybrid A* algorithm is described. The CVAE architecture is introduced and the proposed hybrid algorithm for autonomous parking is explained. 
Section 4 details the verification of the proposed algorithm via simulation. 
Finally, Section 5 summarizes the study and presents the conclusions drawn.

\section{Preliminaries}

\subsection{Hybrid A* Algorithm}

\subsubsection{Transition Model}
\label{subsection:transition_model}
The state $\mathbf{x_k} = (x_k, y_k, \theta_k)^{T}$ represents a state in the planning step $k$, where $x_k$, $y_k$, and $\theta_k$ represent the x-axis position, y-axis position and heading angle, respectively. The state transition model in the discretized form is expressed as follows:
\begin{equation}
\begin{aligned}
x_{k+1} &= x_k + d \cos(\theta_k) dir\\ 
y_{k+1} &= y_k + d \sin(\theta_k) dir\\ 
\theta_{k+1} &= \theta_{k} + \frac{d}{L} \tan(\delta_k) dir
\end{aligned}
\label{eq:transition}
\end{equation}
where $\delta_k$ is a steering angle candidate belonging to a discretized steering angle set $\mathcal{D}$. $dir$ represents the direction of the vehicle motion and $d$ is the expansion amount during one searching step.
The two control actions include the steering angle and direction. $\mathcal{D}$ includes a steering angle set between -40$^{\circ}$ to 40 $^{\circ}$ with an interval of 10$^{\circ}$,
and $dir$ has a value of either -1 or 1, each for backward and forward movement.

\algblockdefx[Foreach]{Foreach}{EndForeach}[1]{\textbf{foreach} #1 \textbf{do}}{\textbf{}}
\IEEEpeerreviewmaketitle
\begin{algorithm}
\caption{Function Hybrid A*}\label{algo:HybridAstar}

\begin{algorithmic}[1]
    \Require {$\mathbf{x_s}$ : start state, $g$ : goal state, $h(x)$ : heuristic, $O$ : obstacle}
    \State $\mathbb{C} \gets \emptyset$
    \State $N = \left\{\mathbf{x_s}, 0, 0, 0, \emptyset \right\} $
    \State $\mathbb{O} \gets N$
    \State $KeepSearching$$\gets$ 1
    \While{$KeepSearching$ is 1}
  
        \If{$\mathbb{O}$ is not empty}
            \State $N$ = \textit{Extract} node with minimum $c+h$ from $\mathbb{O}$ 
            \State \textit{Add} $N$ to $\mathbb{C}$ and \textit{Delete} from $\mathbb{O}$
            \If{$N$ is $g$}
                \State \textit{trajectory} = \textit{Backtracking($N$)}
                \State $KeepSearching$$\gets$ 0
            \Else
                \Foreach {$a \in $Available action($N$)}
                    \State $\left\{\mathbf{x_n},c_{n} \right\} \gets transition(N.x,a,N.c)$
                    \If{\textit{IsCollide}($\mathbf{x_n}$, $O$) is 0} 
                        \State $N_{n} = \left\{\mathbf{x_n},c_{n},h(\mathbf{x_n}),a,N\right\} $
                        \If{$N_{n} \in \mathbb{C}$}
                            \State continue
                        \ElsIf{$N_{n}$ has the same state with \\\hspace{3cm}
                        $n$ $\in$ $\mathbb{O}$ with smaller cost}
                            \State \textit{Replace} $n$ with $N_{n}$
                        \Else
                            \State \textit{Add} $N_{n}$ to $\mathbb{O}$
                        \EndIf
                    \EndIf
                \EndForeach
            \EndIf
        \EndIf
    \EndWhile

\State \Return{$trajectory$}  

\end{algorithmic}

\end{algorithm}

\subsubsection{Hybrid A*}
\label{subsection:hybrid_a_star}
The Hybrid A* algorithm generates a kinodynamic path using the transition model explained in (\ref{eq:transition}). 
The Hybrid A* implementation uses a resolution of 2 m in the X and Y dimensions, and 15$^{\circ}$ in the heading angle. 
This information is used to decide whether the candidate nodes are in a certain grid cell, where only the node with the lowest cost is retained.

\begin{figure*}[t] 
\centering
\includegraphics[width=6.1in]{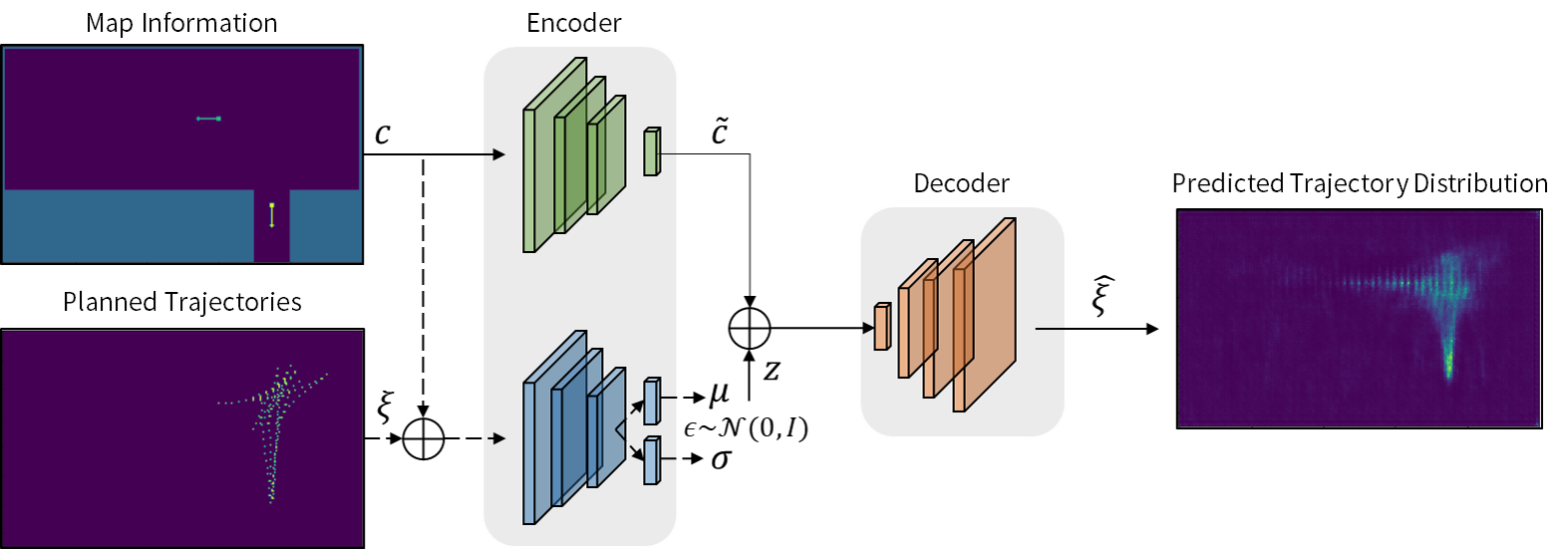}
\caption{Proposed CVAE architecture for neural motion planning.}
\label{fig:overall}
\vspace{-4mm}
\end{figure*}

Algorithm \ref{algo:HybridAstar} shows the procedure that Hybrid A* algorithm follows.
The current state and zero cost are added to the \textit{\textit{open list}}.
The state ($\mathbf{x_k}$), which has the lowest cost, is excluded from the \textit{\textit{open list}} and added to the \textit{closed list}. The search ends if the goal state is reached. If not, the available actions (steering angle and direction) are used to obtain the next state via the transition model. If the searched state ($\mathbf{x_{k+1}}$) is collision-free and is in the \textit{\textit{open list}}, but the sum of the step cost and heuristic is lower than that of the state in the list, the corresponding state is replaced; the state ($\mathbf{x_{k+1}}$) is added to the \textit{open list} if it is not in the \textit{open list}. 
This process is repeated for the states in the \textit{open list}, 
and if a selected state is the goal, the search is finished and the 
optimal trajectory is returned by backtracking via the \textit{closed list}. 
The overall pseudocode is presented in Algorithm 1. $N$ represents a node consisting of $\left\{state, action, path\;cost, heuristic, parent\;node \right\}$. $\mathbb{O}$ and $\mathbb{C}$ indicate the \textit{open list} and \textit{closed list}, respectively. Furthermore, the \textit{IsCollide} procedure returns 0 when the vehicle is collision-free, and the \textit{transition} procedure uses (\ref{eq:transition}) to obtain the next state for the given control action.


\section{Neural Hybrid A* Algorithm}

\subsection{CVAE Architecture}
To learn the distribution of the feasible trajectories for autonomous parking, the CVAE architecture is used, which includes an encoder-decoder network with a conditional input. 
The conditional input utilized in this study represents the map information including the position and heading of the initial and goal points, the obstacles, and free space.

The overall architecture is shown in Fig. \ref{fig:overall}. A two-dimensional (2-D) image is used to represent the map information, $c$. The width and height are 250 and 150, respectively, with a resolution of 0.1 m. Each pixel in the image has a specific value of 0 for free space, 1 for obstacle, 2 for start position with an arrow indicating heading information, 3 for goal position with an arrow indicating heading information.
The true trajectory, $\xi$, is the planned trajectory obtained using the Hybrid A* algorithm. A total of five trajectories are generated and the corresponding pixel in the 2-D image is assigned the value of 1. 
The true trajectory $\xi$, depicted with a dotted line, is used only during the training phase.

The CVAE consists of a generative model $p_{\rho}(\xi|\tilde{c},z)$ and an inference model $q_{\phi}(z|c,\xi)$, and the latent variable $z$ is expressed as follows using the reparameterization trick \cite{kingma2013auto}:
\begin{equation}
z = \mu_{\phi}(c,\xi) + \sigma_{\phi}(c,\xi) \times \epsilon
\label{eq:cvae_z}
\end{equation} 
where $\phi$ and $\rho$ are the parameters of the encoder and decoder networks, respectively. A normal distribution $\mathcal{N}(0,I)$ is utilized to sample $\epsilon$.
To minimize the error between the predicted trajectory distribution $\hat{\xi}$ and the planned trajectory set $\xi$, the reconstruction loss is defined as the $L2$ loss. The CVAE architecture is trained by minimizing the loss function and is defined as follows:

\algblockdefx[Foreach]{Foreach}{EndForeach}[1]{\textbf{foreach} #1 \textbf{do}}{\textbf{}}
\IEEEpeerreviewmaketitle
\begin{algorithm}
\caption{Function Neural Hybrid A*}\label{algo:NeuralHybridAstar}

\begin{algorithmic}[1]
    \Require {$\mathbf{x_s}$ : start state, $g$ : goal state, $h(x)$ : heuristic, $O$ : obstacle, $Dmap$ : predicted trajectory distribution map}
    \State $\mathbb{C} \gets \emptyset$
    \State $N = \left\{\mathbf{x_s}, 0, 0, 0, \emptyset \right\} $
    \State $\mathbb{O} \gets N$
    \State $KeepSearching$$\gets$ 1
    \While{$KeepSearching$ is 1}
  
        \If{$\mathbb{O}$ is not empty}
            \State $N$ = \textit{Extract} node with minimum $c+h$ from $\mathbb{O}$ 
            \State \textit{Add} $N$ to $\mathbb{C}$ and \textit{Delete} from $\mathbb{O}$
            \If{$N$ is $g$}
                \State \textit{trajectory} = \textit{Backtracking($N$)}
                \State $KeepSearching$$\gets$ 0
            \Else
                \Foreach {$a \in $Available action($N$)}
                    \State $\left\{\mathbf{x_n},c_{n} \right\} \gets transition(N.x,a,N.c)$
                    \If{\textit{Rand()} $>$ 0.2}
                        \If{\textit{CheckDistMap}$(\mathbf{x_n},Dmap)$}
                            \State continue
                        \EndIf
                    \EndIf
                    \If{\textit{IsCollide}($\mathbf{x_n}$, $O$) is 0} 
                        \State $N_{n} = \left\{\mathbf{x_n},c_{n},h(\mathbf{x_n}),a,N\right\} $
                        \If{$N_{n} \in \mathbb{C}$}
                            \State continue
                        \ElsIf{$N_{n}$ has the same state with \\\hspace{3cm}
                        $n$ $\in$ $\mathbb{O}$ with smaller cost}
                            \State \textit{Replace} $n$ with $N_{n}$
                        \Else
                            \State \textit{Add} $N_{n}$ to $\mathbb{O}$
                        \EndIf
                    \EndIf
                \EndForeach
            \EndIf
        \EndIf
    \EndWhile

\State \Return{$trajectory$}  

\end{algorithmic}

\end{algorithm}

\DeclarePairedDelimiterX{\norm}[1]{\lVert}{\rVert}{#1}
\begin{equation}
\begin{split}
\mathcal{L} = &L_{REC} + L_{KL} \\
= &\norm{\xi - \hat{\xi}}^2 + \beta D_{KL}(q_{\phi}(z|c,\xi)||p(z))
\label{eq:cvae_loss}
\end{split}
\end{equation} 
where the hyperparameter $\beta$ balances the two losses. The former loss represents the reconstruction loss, and the latter is the KL divergence loss between the multivariate normal distribution and the output distribution from the encoder.

The true trajectory set $\xi$ is not available in the test phase. Therefore, $z$ is directly sampled from $\mathcal{N}(0,I)$, and only the decoder part is utilized. Condition $c$ is passed through an encoder to generate a latent vector $\tilde{c}$, which is used in the decoder for inference.

\begin{figure*}[t] 
\centering
\includegraphics[width=6.5in]{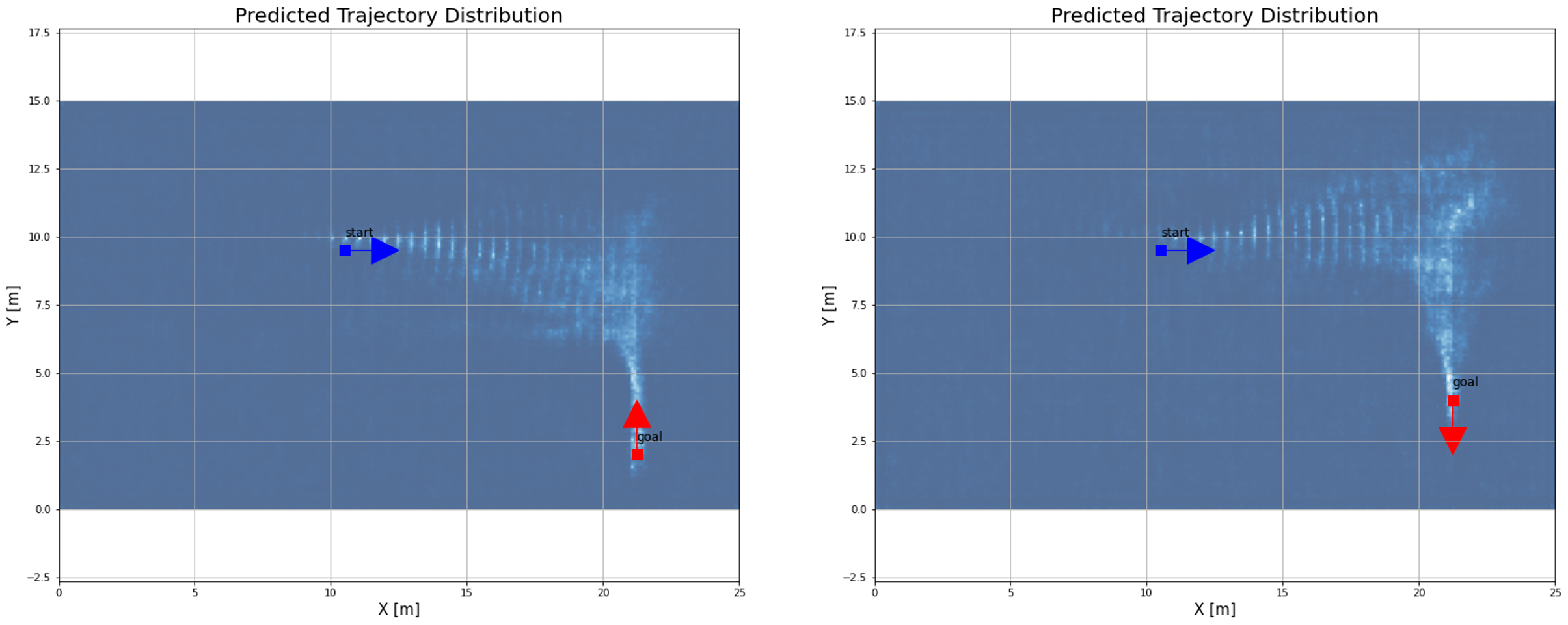}
\caption{Example of predicted trajectory distribution: Results (a) with the heading angle of 90$^{\circ}$ at goal position; (b) with the heading angle of -90$^{\circ}$ at goal position.}
\label{fig:compare_heading}

\end{figure*}

Various predicted trajectory distributions can be generated by feeding different conditions such as map information.
For example, by varying the start and goal positions along with the heading value, different trajectory distributions are generated. Fig. \ref{fig:compare_heading} shows the predicted trajectory distribution when the heading angle of the goal position is the opposite while the starting position and the target parking space are the same. The result indicates that the proposed CVAE network is successfully trained and can distinguish the difference in the map information.

\begin{figure}[h] 
\centering
\includegraphics[width=3.0in]{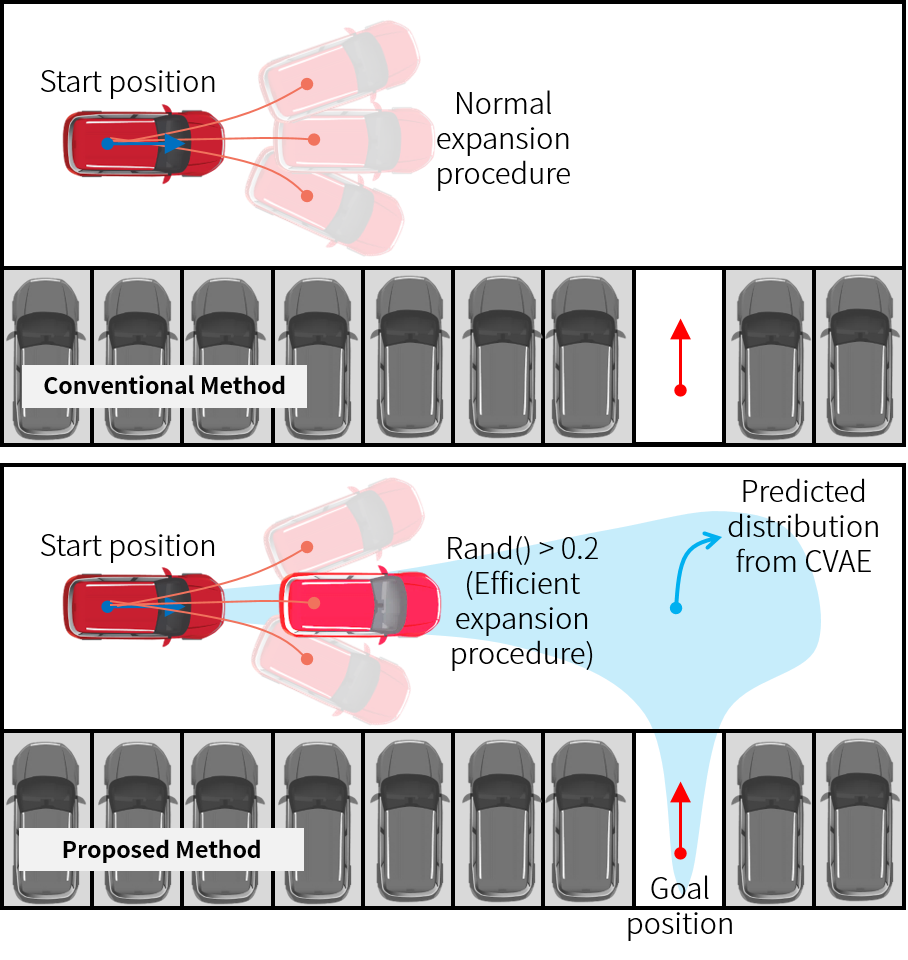}
\caption{Example of the proposed planning strategy based on the predicted distribution.}
\label{fig:comparison}
\end{figure}

\subsection{Neural Hybrid A*}
Neural Hybrid A* uses trajectory distribution map information obtained from the CVAE network. The details are presented in Algorithm \ref{algo:NeuralHybridAstar} where the requirement includes new input, and Lines 15 to 17 are newly added. 
$Dmap$ represents the distribution map of the predicted trajectory.
The Line 15, a random number $Rand() \in (0,1)$ is checked to determine whether $Dmap$ is utilized. In this study, 80$\%$ of the node expansion proceeds with the help of a neural network model. 
The $CheckDistMap$ procedure returns 0 if the current state in the $Dmap$ has a value greater than a certain threshold. 
As seen in Lines 16 and 17, if the $Dmap$ has a value below the threshold in the position of state $\mathbf{x_n}$, the action chosen in Line 13 is deemed unnecessary; otherwise, the next step is to proceed. 
Thus, with this state expansion strategy, the control action is chosen efficiently by performing the expansion mostly based on the learned trajectory distribution.
Fig. \ref{fig:comparison} represents an example image of the aforementioned planning strategy. The expansion by the predefined action set is performed in the upper image which uses the conventional method. In the bottom image, an efficient expansion method is performed based on the predicted distribution from CVAE.

\section{Simulation}
In this section, the dataset acquisition, implementation details, evaluation of the CVAE model, and the results are discussed. 
The neural network model was designed and trained using PyTorch \cite{paszke2019pytorch}, an open source Python library. Hybrid A*, a motion planning algorithm, is implemented in Python.

\begin{figure*}[t] 
\centering
\includegraphics[width=6.5in]{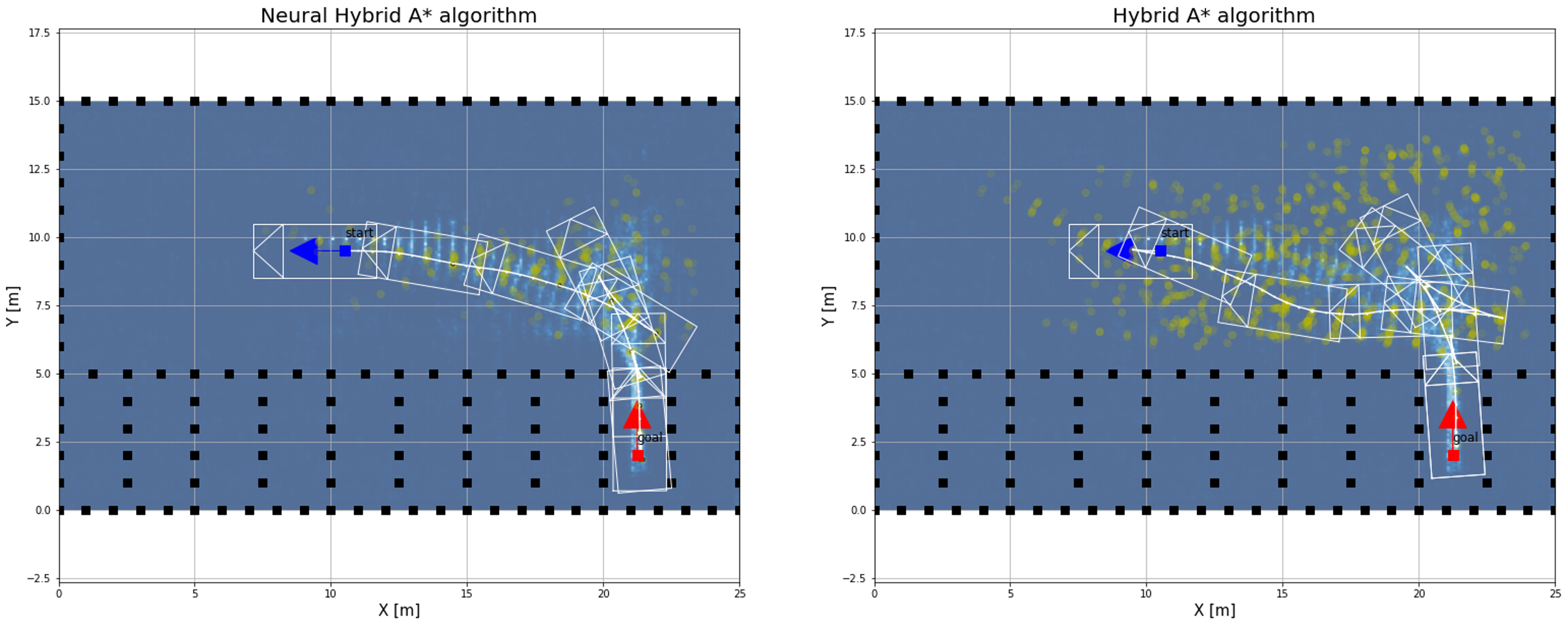}
\caption{Comparison between the proposed method and the conventional method in scenario 1. The blue distribution represents the predicted distribution of the feasible trajectory from the proposed CVAE network. The yellow points indicate the expanded nodes during the planning process.}
\label{fig:result1_one_traj}
\end{figure*}

\begin{figure*}[t] 
\centering
\includegraphics[width=6.5in]{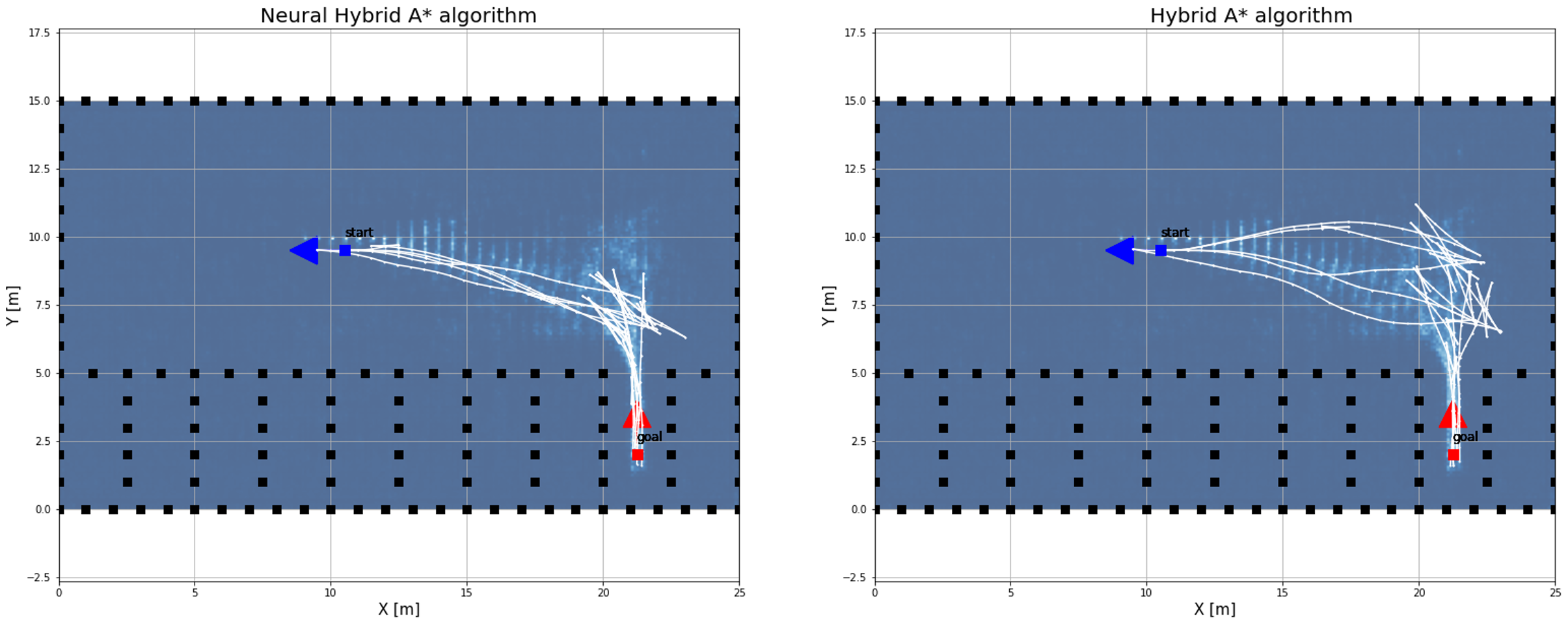}
\caption{Five trajectories generated in scenario 1: Results (a) with the proposed method; (b) with the conventional method.}
\label{fig:result1_compare}
\end{figure*}

\subsection{Dataset} 
\label{sub:dataset}
To train the model, expert trajectory sets are collected using the Hybrid A* algorithm. Each scene has a combination of five trajectories. 
The hybrid A* algorithm is a deterministic method, resulting in the same solution.
Therefore, the order of action set $\mathcal{D}$ is randomly mixed to obtain different trajectories for the proposed network training.
A total of 1,000 scenarios are simulated for the data collection. The start position is randomly sampled in free space, and the heading angle is sampled from either 0$^{\circ}$ or 180$^{\circ}$.
The parking space is randomly selected, and the goal position is selected accordingly. The heading angle of the goal position is randomly sampled either from -90$^{\circ}$ or 90$^{\circ}$. The collected dataset is utilized as a conditional input and label for training the proposed CVAE network.

\subsection{Implementation Details}
\label{sub:implemetation_details}
The overall network is composed of an encoder and a decoder. There are two encoders, one for encoding the map information to be used in the decoder, and the other is for encoding the planned trajectories along with the map information to produce a latent vector.

The encoder is composed of a 2-D convolution, batch normalization, rectified linear unit (ReLU) activation function, and fully-connected layers.
The convolutional neural network (CNN) model consists of three convolutional layers with output channels of 16, 32 and 64. The kernel size is [4,4], and the stride is 2.
For the first encoder, the result is passed through a fully-connected layer with 32 hidden units, which is the dimension of the encoded condition.
For the second encoder, the result is passed through two fully-connected layers with 32 hidden units, which is the dimension of the latent variable.

The decoder is composed of a fully-connected layer, 2-D deconvolution, batch normalization, and ReLU. 
The CNN model for the decoder consists of three convolutional layers with output channels of 32, 16 and 1. The kernel size and stride are the same as those of the encoder.

In the training process, the optimization was performed using a standard Adam optimizer with a learning rate of 0.001. In addition, parameter $\beta$ for the CVAE loss is set to 0.1.

\subsection{Evaluation of the CVAE Model} 
\label{sub:cvae_eval}
\begin{figure*}[t] 
\centering
\includegraphics[width=6.5in]{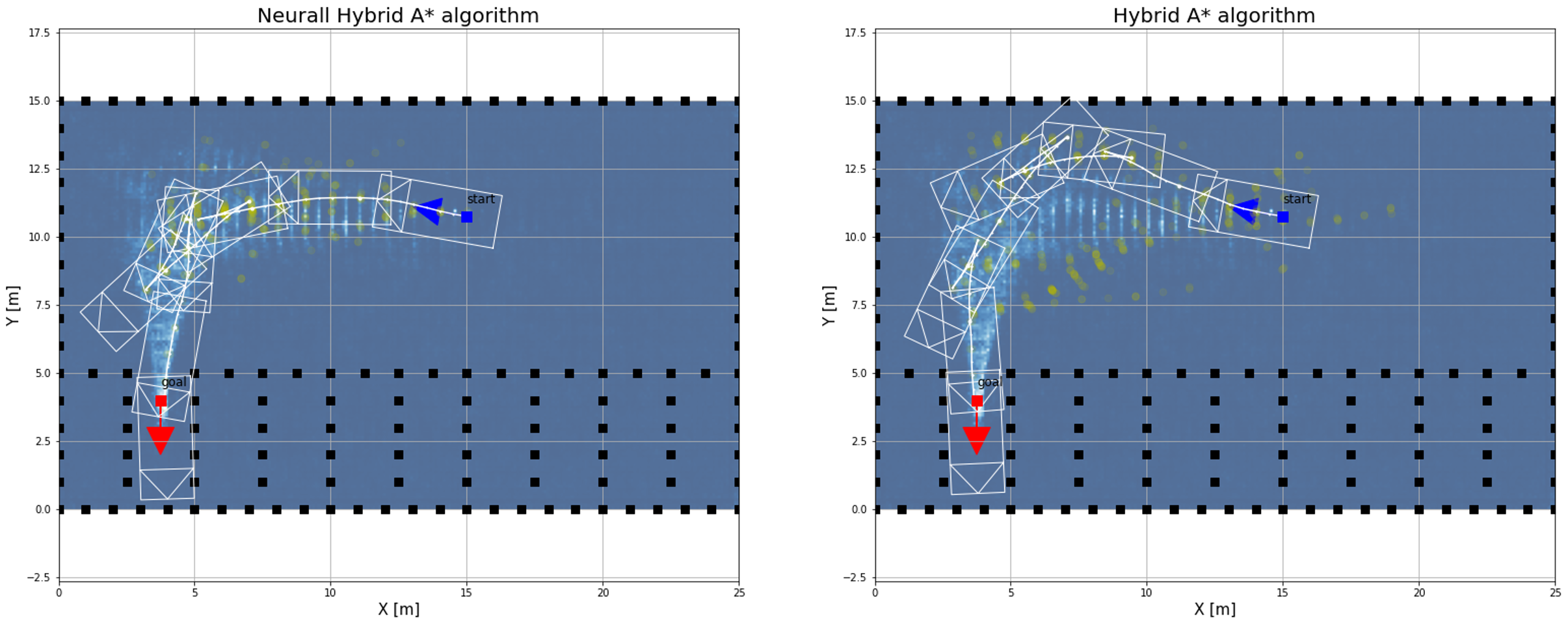}
\caption{Comparison between the proposed method and the conventional method in scenario 2.}
\label{fig:result2_one_traj}
\end{figure*}

\begin{figure*}[t] 
\centering
\includegraphics[width=6.5in]{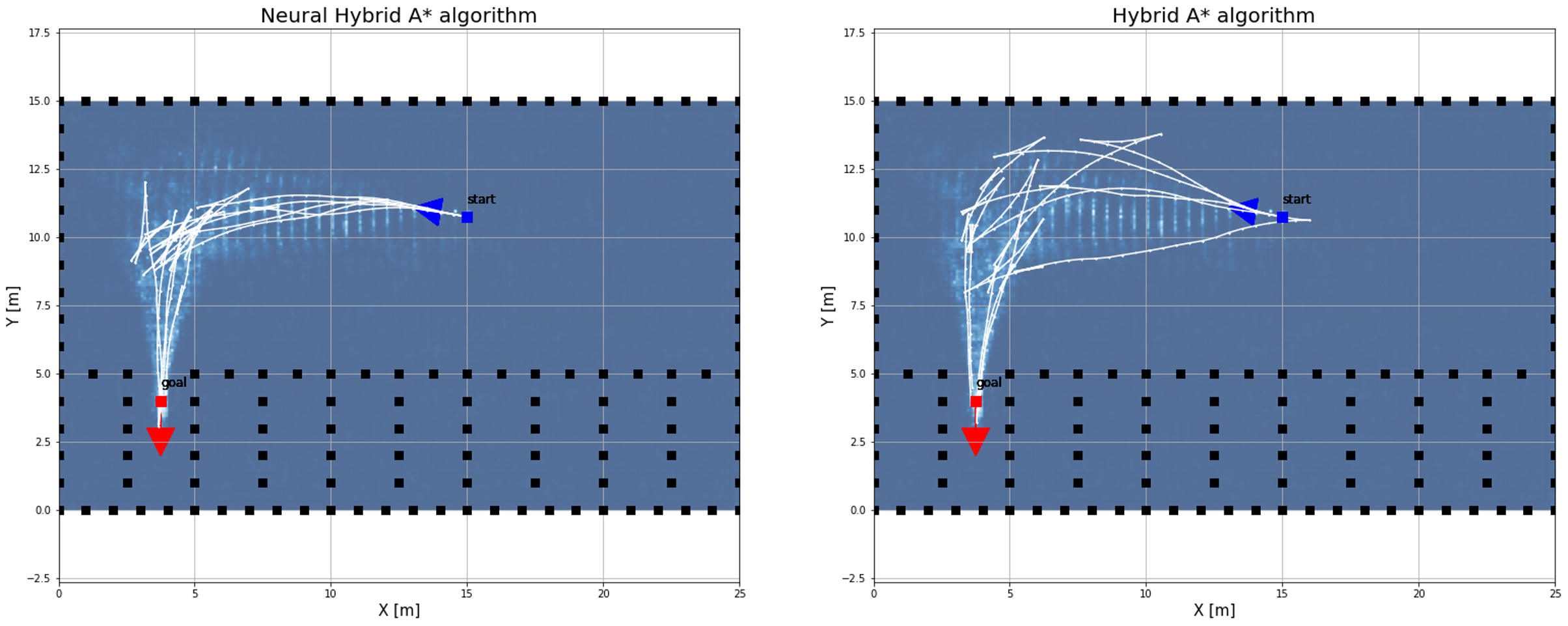}
\caption{Five trajectories generated in scenario 2: Results (a) with the proposed method; (b) with the conventional method.}
\label{fig:result2_compare}
\end{figure*}

\begin{figure*}[t] 
\centering
\includegraphics[width=6.5in]{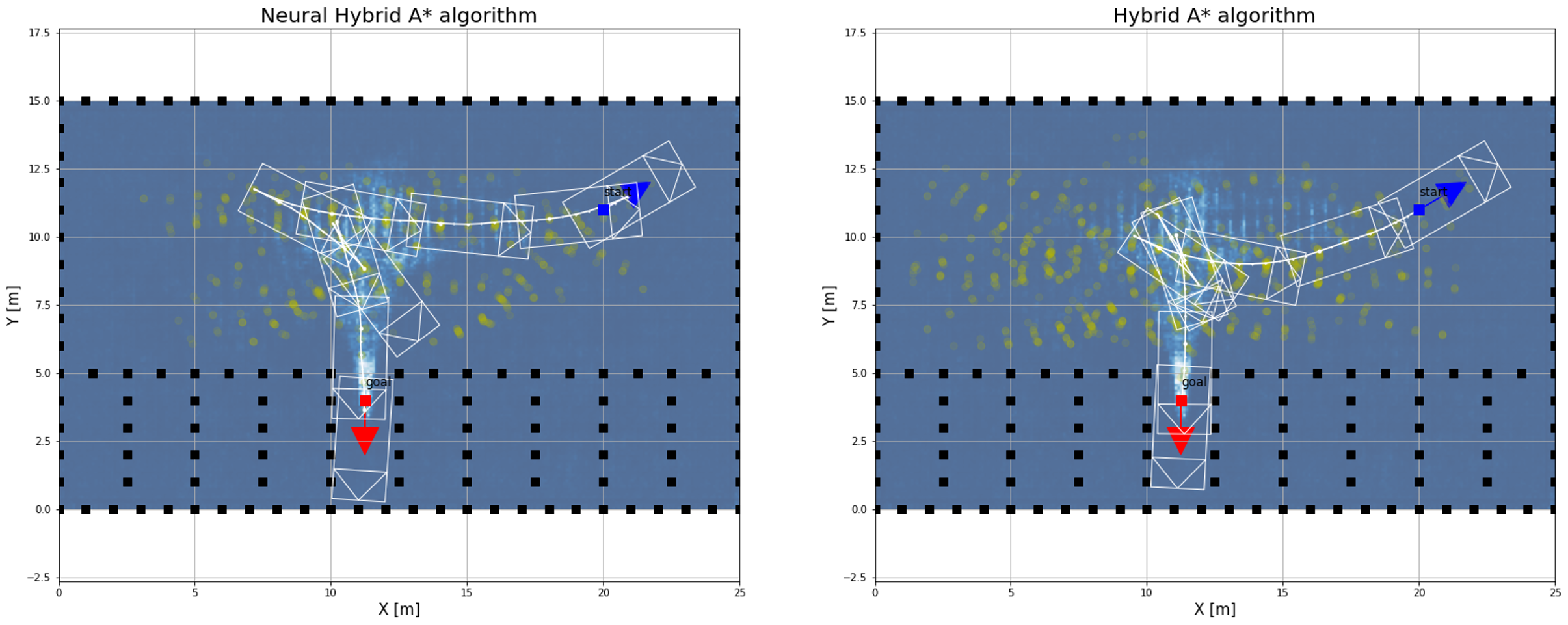}
\caption{Comparison between the proposed method and the conventional method in scenario 3.}
\label{fig:result3_one_traj}
\end{figure*}

\begin{figure*}[t] 
\centering
\includegraphics[width=6.5in]{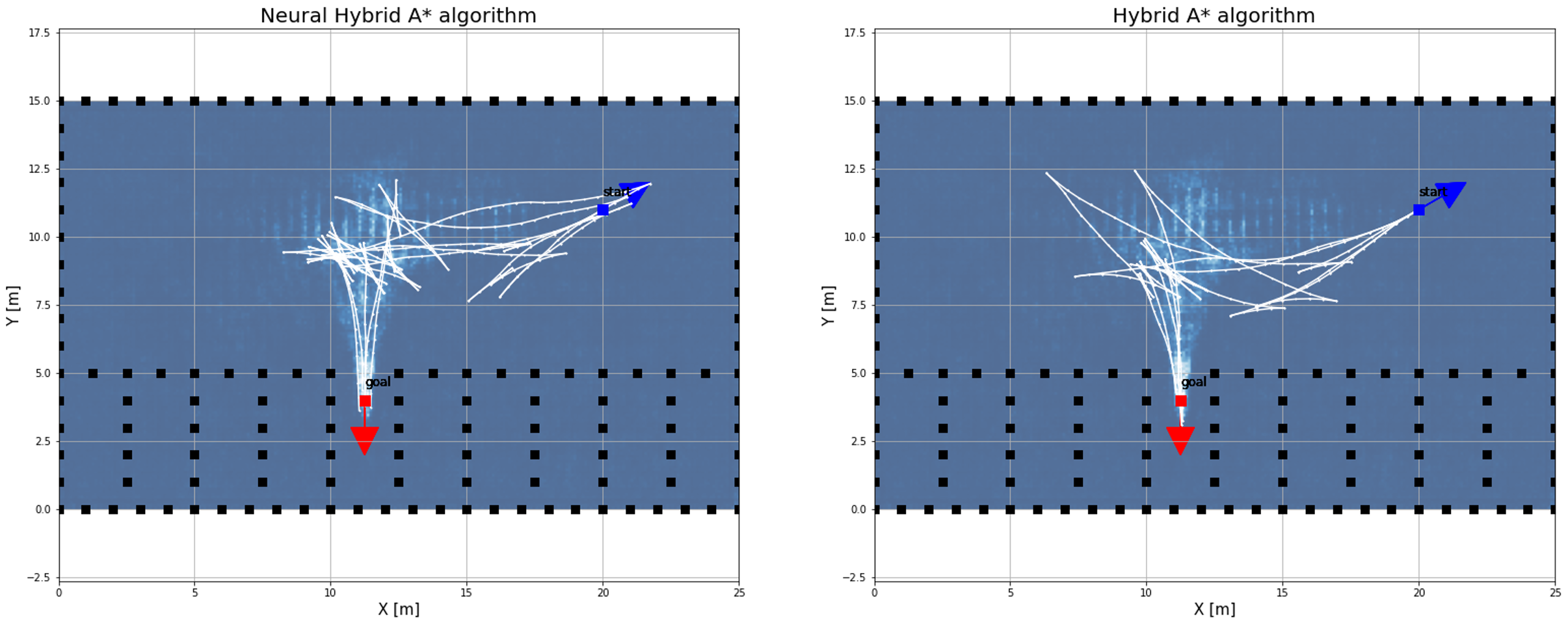}
\caption{Five trajectories generated in scenario 3: Results (a) with the proposed method; (b) with the conventional method.}
\label{fig:result3_compare}
\end{figure*}

\begin{figure*}[t] 
\centering
\includegraphics[width=6.5in]{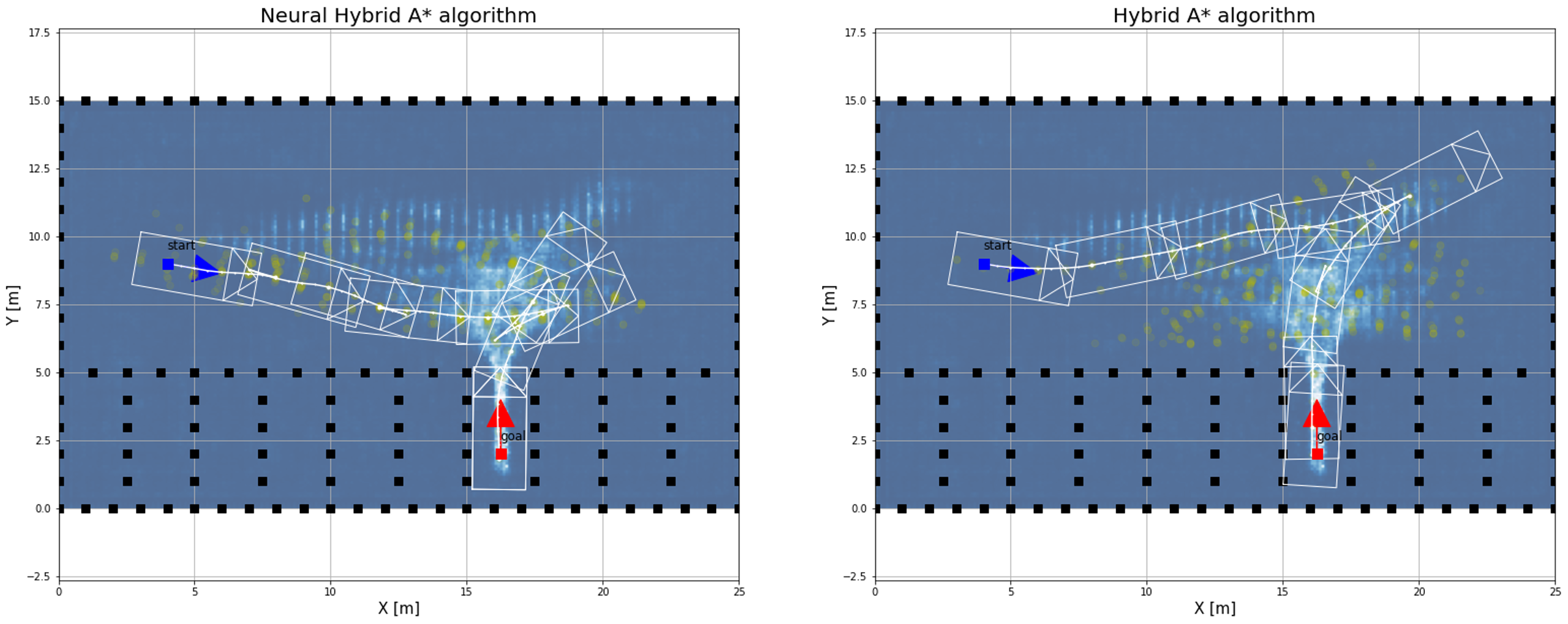}
\caption{Comparison between the proposed method and the conventional method in scenario 4.}
\label{fig:result4_one_traj}
\end{figure*}

To evaluate the performance of the proposed method, a test dataset is generated. The map information with different initial positions or heading angles with those of the training set is selected for inference. 
The Figs. \ref{fig:result1_compare} to \ref{fig:result4_one_traj} show the results of the proposed neural Hybrid A* algorithm compared with that of the conventional Hybrid A* algorithm.

In Fig. \ref{fig:result1_one_traj}, the results for the first scenario are compared. The blue distribution represents the predicted distribution of the feasible trajectory from the proposed CVAE network. The yellow circles represent the expanded nodes during this process.
The left figure shows the results of the proposed method. As can be seen, along with the expanded nodes, the expansion is mostly done on the learned distribution. 
In contrast, the right figure shows the result without the help of the learned distribution.
The nodes are scattered regardless of the distribution, resulting in high computation.

Fig. \ref{fig:result1_compare} shows the planning result with the five trajectories generated. As shown in the figure on the left, the generated trajectories are mostly on the predicted distribution. 
The right figure shows that the planning results are more random regardless of the distribution, as no information on the CVAE network is used.

Fig. \ref{fig:result2_one_traj} shows the results for the second scenario. The heading angle of the initial point is set to 170$^{\circ}$, which is not included in the training dataset. The generated distribution seems to be utilized well as a guidance map as the expansion is nearly completed on the distribution map. However, as seen in the right figure, the expansion proceeds by covering a wider range when the distribution map is not utilized, resulting in a higher cost of node expansion.

Fig. \ref{fig:result2_compare} also shows the results with the five trajectories generated. As shown in the left figure, the generated trajectories are mostly on the predicted distribution using the neural Hybrid A* method. However, it show more random results without the distribution map.

\begin{figure*}[t] 
\centering
\includegraphics[width=6.5in]{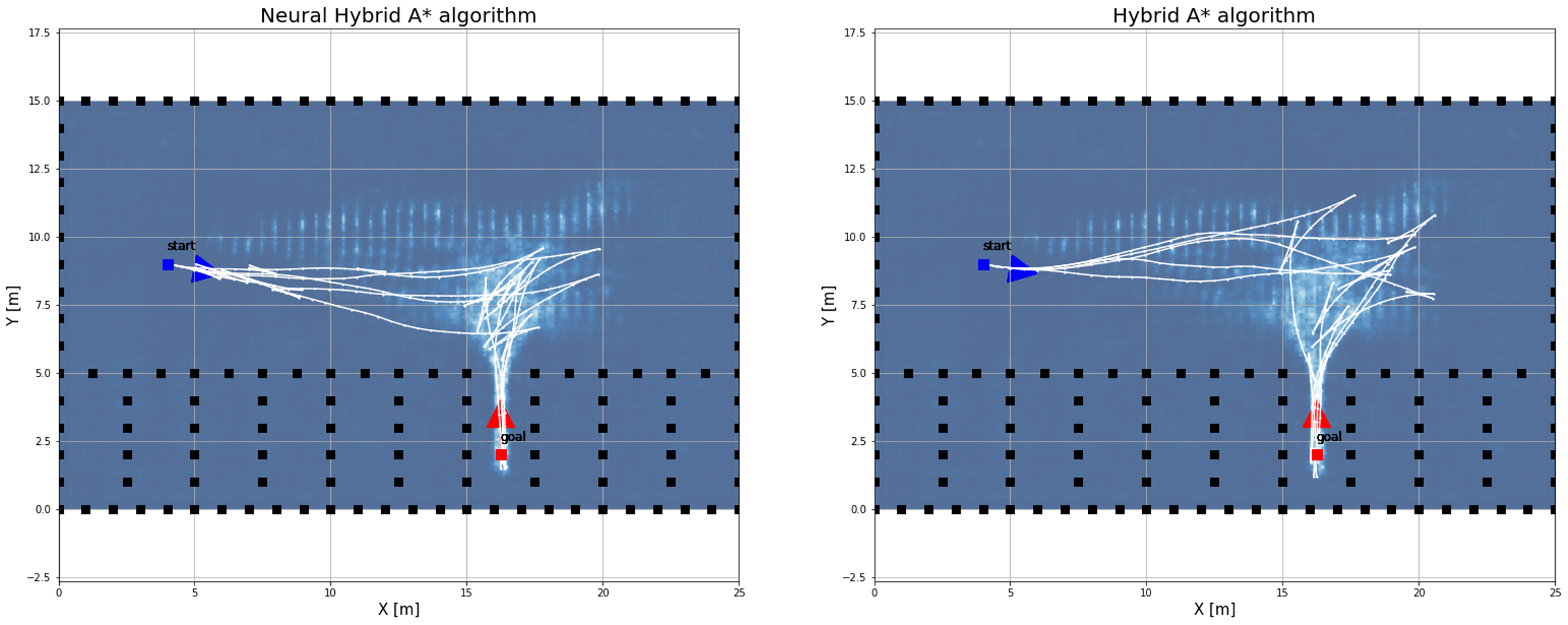}
\caption{Five trajectories generated in scenario 4: Results (a) with the proposed method; (b) with the conventional method.}
\label{fig:result4_compare}
\end{figure*}

Figs. \ref{fig:result3_one_traj} to \ref{fig:result4_compare} show similar results to those of the aforementioned scenarios.
In the third scenario shown in Fig. \ref{fig:result3_compare}, the CVAE network provides a feasible distribution map even with an initial heading angle of 30$^{\circ}$ which varies significantly from the training dataset. 

\begin{table}[h]
\caption{Comparison of metrics.}
\setlength{\tabcolsep}{1.2pt}
\label{TABLE:compare}
\begin{center}
\begin{tabular}{|c|c|c|c|c|c|} 
\hline 
& & \small Scen. 1 & \small Scen. 2 & \small Scen. 3 & \small Scen. 4 \\
\hline 
\small Comparison & \small Time & \small 24.75\%  & \small 30.33\% & \small 50.0\%  &  \small 10.86\%\\
\small (vs Hybrid A*) & \small Node & \small 50.91\%  & \small 53.30\% & \small 79.36\%  & \small 35.35\% \\
\hline 
\end{tabular}
\end{center}
\end{table}

To evaluate the performance of the neural Hybrid A* algorithm, two metrics, which comprise computational time and the number of nodes in the \textit{open list}, are evaluated under the four test scenarios.
Table. \ref{TABLE:compare} shows the performance comparison in terms of the metrics. 
A statistical result is obtained by evaluating the scenarios five times and using a mean value for the two metrics.
In all the test scenarios, the computational time and the number of nodes are significantly reduced with the proposed neural Hybrid A* method compared with the conventional Hybrid A* algorithm.
In terms of the computational time, the results illustrate that the neural Hybrid A* algorithm improves the state expansion efficiency of the planning problem for autonomous parking. 
In addition, in terms of the node, the results indicate that the proposed method is much less memory-intensive than the conventional method.

\section{Conclusions}
This paper presents a novel hybrid motion planning strategy for autonomous parking that integrates a deep generative network and a conventional searching based path planning algorithm.
The CVAE network is constructed to learn the predicted distribution of the feasible trajectories, given the initial and goal positions along with the obstacle information.
The proposed model learns several maps with feasible paths generated by the Hybrid A* algorithm. 

The simulation results with the neural Hybrid A* algorithm shows that 
the neural Hybrid A* algorithm reduces the computational time and number of nodes significantly.
As the proposed method uses an expansion strategy based on the predicted distribution, other searching-based planning algorithms can also be applied.
In future studies, simulations in more diverse parking scenarios will be conducted.

\bibliographystyle{IEEEtran}
\bibliography{main}

\begin{thebibliography}{10}
\providecommand{\url}[1]{#1}
\csname url@rmstyle\endcsname
\providecommand{\newblock}{\relax}
\providecommand{\bibinfo}[2]{#2}
\providecommand\BIBentrySTDinterwordspacing{\spaceskip=0pt\relax}
\providecommand\BIBentryALTinterwordstretchfactor{4}
\providecommand\BIBentryALTinterwordspacing{\spaceskip=\fontdimen2\font plus
\BIBentryALTinterwordstretchfactor\fontdimen3\font minus
  \fontdimen4\font\relax}
\providecommand\BIBforeignlanguage[2]{{%
\expandafter\ifx\csname l@#1\endcsname\relax
\typeout{** WARNING: IEEEtran.bst: No hyphenation pattern has been}%
\typeout{** loaded for the language `#1'. Using the pattern for}%
\typeout{** the default language instead.}%
\else
\language=\csname l@#1\endcsname
\fi
#2}}

\bibitem{khalid2020smart}
M.~Khalid, K.~Wang, N.~Aslam, Y.~Cao, N.~Ahmad, and M.~K. Khan, ``From smart
  parking towards autonomous valet parking: A survey, challenges and future
  works,'' \emph{Journal of Network and Computer Applications}, p. 102935,
  2020.

\bibitem{dong2016experimental}
Y.~Dong, Y.~Zhang, and J.~Ai, ``Experimental test of artificial potential
  field-based automobiles automated perpendicular parking,''
  \emph{International Journal of Vehicular Technology}, vol. 2016, 2016.

\bibitem{kuwata2008motion}
Y.~Kuwata, G.~A. Fiore, J.~Teo, E.~Frazzoli, and J.~P. How, ``Motion planning
  for urban driving using rrt,'' in \emph{2008 IEEE/RSJ International
  Conference on Intelligent Robots and Systems}.\hskip 1em plus 0.5em minus
  0.4em\relax IEEE, 2008, pp. 1681--1686.

\bibitem{vlasak2020accelerated}
J.~Vlasak, M.~Sojka, and Z.~Hanz{\'a}lek, ``Accelerated rrt* and its evaluation
  on autonomous parking,'' \emph{arXiv preprint arXiv:2002.04521}, 2020.

\bibitem{benenson2006integrating}
R.~Benenson, S.~Petti, T.~Fraichard, and M.~Parent, ``Integrating perception
  and planning for autonomous navigation of urban vehicles,'' in \emph{2006
  IEEE/RSJ International Conference on Intelligent Robots and Systems}.\hskip
  1em plus 0.5em minus 0.4em\relax IEEE, 2006, pp. 98--104.

\bibitem{cheng2014improved}
L.~Cheng, C.~Liu, and B.~Yan, ``Improved hierarchical a-star algorithm for
  optimal parking path planning of the large parking lot,'' in \emph{2014 IEEE
  International Conference on Information and Automation (ICIA)}.\hskip 1em
  plus 0.5em minus 0.4em\relax IEEE, 2014, pp. 695--698.

\bibitem{dolgov2010path}
D.~Dolgov, S.~Thrun, M.~Montemerlo, and J.~Diebel, ``Path planning for
  autonomous vehicles in unknown semi-structured environments,'' \emph{The
  international journal of robotics research}, vol.~29, no.~5, pp. 485--501,
  2010.

\bibitem{sharma2021recent}
O.~Sharma, N.~C. Sahoo, and N.~Puhan, ``Recent advances in motion and behavior
  planning techniques for software architecture of autonomous vehicles: A
  state-of-the-art survey,'' \emph{Engineering Applications of Artificial
  Intelligence}, vol. 101, p. 104211, 2021.

\bibitem{bicchi2002reachability}
A.~Bicchi, A.~Marigo, and B.~Piccoli, ``On the reachability of quantized
  control systems,'' \emph{IEEE Transactions on Automatic Control}, vol.~47,
  no.~4, pp. 546--563, 2002.

\bibitem{pivtoraiko2005efficient}
M.~Pivtoraiko and A.~Kelly, ``Efficient constrained path planning via search in
  state lattices,'' in \emph{International Symposium on Artificial
  Intelligence, Robotics, and Automation in Space}.\hskip 1em plus 0.5em minus
  0.4em\relax Munich Germany, 2005, pp. 1--7.

\bibitem{ichter2018learning}
B.~Ichter, J.~Harrison, and M.~Pavone, ``Learning sampling distributions for
  robot motion planning,'' in \emph{2018 IEEE International Conference on
  Robotics and Automation (ICRA)}.\hskip 1em plus 0.5em minus 0.4em\relax IEEE,
  2018, pp. 7087--7094.

\bibitem{wang2020neural}
J.~Wang, W.~Chi, C.~Li, C.~Wang, and M.~Q.-H. Meng, ``Neural rrt*:
  Learning-based optimal path planning,'' \emph{IEEE Transactions on Automation
  Science and Engineering}, vol.~17, no.~4, pp. 1748--1758, 2020.

\bibitem{chen2016dynamic}
N.~Chen, M.~Karl, and P.~Van Der~Smagt, ``Dynamic movement primitives in latent
  space of time-dependent variational autoencoders,'' in \emph{2016 IEEE-RAS
  16th international conference on humanoid robots (Humanoids)}.\hskip 1em plus
  0.5em minus 0.4em\relax IEEE, 2016, pp. 629--636.

\bibitem{sohn2015learning}
K.~Sohn, H.~Lee, and X.~Yan, ``Learning structured output representation using
  deep conditional generative models,'' \emph{Advances in neural information
  processing systems}, vol.~28, pp. 3483--3491, 2015.

\bibitem{feng2019vehicle}
X.~Feng, Z.~Cen, J.~Hu, and Y.~Zhang, ``Vehicle trajectory prediction using
  intention-based conditional variational autoencoder,'' in \emph{2019 IEEE
  Intelligent Transportation Systems Conference (ITSC)}.\hskip 1em plus 0.5em
  minus 0.4em\relax IEEE, 2019, pp. 3514--3519.

\bibitem{bhattacharyya2019conditional}
A.~Bhattacharyya, M.~Hanselmann, M.~Fritz, B.~Schiele, and C.-N. Straehle,
  ``Conditional flow variational autoencoders for structured sequence
  prediction,'' \emph{arXiv preprint arXiv:1908.09008}, 2019.

\bibitem{kingma2013auto}
D.~P. Kingma and M.~Welling, ``Auto-encoding variational bayes,'' \emph{arXiv
  preprint arXiv:1312.6114}, 2013.

\bibitem{paszke2019pytorch}
A.~Paszke, S.~Gross, F.~Massa, A.~Lerer, J.~Bradbury, G.~Chanan, T.~Killeen,
  Z.~Lin, N.~Gimelshein, L.~Antiga, \emph{et~al.}, ``Pytorch: An imperative
  style, high-performance deep learning library,'' \emph{Advances in neural
  information processing systems}, vol.~32, pp. 8026--8037, 2019.

\end{thebibliography}

\end{document}